\documentclass{article}

\usepackage{times}
\usepackage{graphicx} 
\usepackage{subfigure} 
\usepackage{comment}
\usepackage{multirow}

\usepackage{natbib}

\usepackage{algorithm}
\usepackage{algorithmic}

\usepackage{hyperref}

\usepackage[accepted]{icml2017}

\icmltitlerunning{Detecting early signs of depressive and manic episodes in patients with bipolar disorder using the signature-based model}

\begin{document} 

\twocolumn[
\icmltitle{Detecting early signs of depressive and manic episodes in patients \\ with bipolar disorder using the signature-based model}




\begin{icmlauthorlist}
\icmlauthor{Andrey Kormilitzin}{mi}
\icmlauthor{Kate E.A. Saunders}{dp,ohnhsft}
\icmlauthor{Paul J. Harrison}{dp,ohnhsft}
\icmlauthor{John R. Geddes}{dp,ohnhsft}
\icmlauthor{Terry Lyons}{mi}
\end{icmlauthorlist}

\icmlaffiliation{mi}{Mathematical Institute, University of Oxford, Andrew Wiles Building, Woodstock Rd, Oxford OX2 6GG, UK}
\icmlaffiliation{dp}{Department of Psychiatry, University of Oxford}
\icmlaffiliation{ohnhsft}{Oxford Health NHS Foundation Trust, Warneford Hospital,Oxford OX3 7JX, UK }

\icmlcorrespondingauthor{}{andrey.kormilitzin@maths.ox.ac.uk}

\icmlkeywords{stochastic analysis, sequential data, classification, bipolar disorder, digital healthcare}

\vskip 0.3in
]



\printAffiliationsAndNotice{} 

\begin{abstract} 
\textbf{Background:} Recurrent major mood episodes and subsyndromal mood instability cause substantial disability in patients with bipolar disorder. Early identification of mood episodes enabling timely mood stabilisation is an important clinical goal. Recent technological advances allow the prospective reporting of mood in real time enabling more accurate, efficient data capture. The complex nature of these data streams in combination with challenge of deriving meaning from missing data mean pose a significant analytic challenge. The signature method is derived from stochastic analysis and has the ability to capture important properties of complex ordered time series data.  

\textbf{Objective:} To explore whether the onset of episodes of mania and depression can be identified using self-reported mood data.

\textbf{Methods:} Self-reported mood data were collected from 261 participants with bipolar disorder using the True Colours monitoring system. Manic and depressive episodes were defined as an ASRM score ≥ 6 or a QIDS score ≥ 10, respectively. The signature method was used to extract features from a rolling window of k-weeks, where k = $\{$4, 6, 8, 12, 20, 50$\}$. Two independent models for prediction of depressive and manic mood episodes were trained using logistic regression with the elastic net regularisation scheme to reduce overfitting. The stability of the generalisation error metrics of the predictive models was estimated by pooling 100 repetitions of 10-fold cross validation. 

\textbf{Results:} The signature method on average accurately predicted 79.2$\%$ of precursors to depressive episodes with a sensitivity of 76.9$\%$ and specificity of 79.5$\%$ (PPV=78.6$\%$, AUC=0.86) and 71.9$\%$ of precursors to manic episodes with a sensitivity of 73.3$\%$ and specificity 79.2$\%$ (PPV=77.1$\%$, AUC=0.83). This was more accurate than predictions (accuracy 77.4$\%$ for depression and 69.5$\%$ for mania, p-value $<$ 0.001) based up on a model comprises three features: the mean value, variability and a number of missing responses within the window.

\textbf{Conclusions:} The signature method offers a systematic approach to the analysis of longitudinal self-reported mood data. The accuracies of the signature-based linear models are considerably better than linear models based on manually crafted features and the models have the potential to significantly enhance self-management and clinical care in bipolar disorder.
\end{abstract} 

\section{Introduction}
\label{intro}

\subsection{Related Work}

Predicting the future outcomes on the basis of historical observations is a long standing challenge in many scientific areas. The accurate prediction of future symptoms in patients with bipolar disorder could facilitate timely clinical interventions enhancing patients’ quality of life, altering the course of disease and could have economic benefits for the national health service \cite{manning2005burden}. Several predictive models have been proposed using the self-reported longitudinal mood data from psychometric questionnaires. These models included: autoregressive linear models \cite{moore2012forecasting, moore2014mood}, relaxation oscillator framework \cite{bonsall2015bipolar}, Kalman filter with the data from geographic location \cite{tsanas2016daily, palmius2017detecting} and an effective subgroup selection among diverse cohort of patients \cite{palmius2016nonparametrics}. However, these methods focused only on predicting the numerical value of the rating scale In this work, we took a different route in building a predictive model. In contrast to previous approaches, we aim to identify the precursors of upcoming episodes of depression and mania, using the self-reported rating scale. In fact, we solve a binary classification problem, where our developed model, using an interval of k-consecutive self-reported observation, estimates a probability of being a precursor to an episode. 

\section{Methods} 
 
\subsection{Data}

The data were collected as part of the OXTEXT-1 study \cite{bilderbeck2017effects}.  The participants completed standardised questionnaires on a weekly basis using the True Colours mood monitoring system after receiving a text or email prompt. The data were collected in an observational manner and independent from the clinical care.  

Self-reported mood data using the Quick Inventory of Depressive Symptoms (QIDS-SR16) \cite{rush200316} and Altman Self-Rating Mania scale (ASRM) \cite{altman1997altman}.  A depressive episode was defined as a QIDS score of above 11 for at least two consecutive weeks. A manic episode was defined as an AMRS score above a threshold 6 for at least one week \cite{kessler2005lifetime}.  

\subsection{Demographics and Patient Selection}
\label{demogr}

The original cohort contained 286 subjects. Data collected from 01-Jan-2012 until 31-Dec-2016 was included in this analysis. We excluded 25 participants from the analysis: 22 participants had unconfirmed diagnoses and 3 participants withdrew consent. Of the 261 included participants 148 and 113 subjects were diagnosed with bipolar type I and type II, (denoted by BP-I and BP-II) respectively. All identical duplicate values were removed. If there were multiple responses within a week, only the first response was considered.

Additionally, we excluded patients who stayed less than 5 weeks in self-monitoring and all of those who had no recorded mood episodes during the study period. From the exploratory analysis, we found that only 59$\%$ of patients (155/261) in this cohort had episodes of depression (spent at least two weeks in depression) and 69$\%$ (181/261) had manic episodes (at least one week).  The reported ethnicity and education level in accordance with the standard classification. One male from the BP-II group did not report his ethnicity. One male from the BP-I group, one male from the BP-II group, two females, one BP-I and one BP-II did not report their education levels. Percents of relevant groups may not add up to 100 due to rounding errors. The adjusted adherence corresponds to the period of observations between the first and the last actual responses. The demographic data summarised in Table \ref{table:demographics_results}.

\subsection{Signature-based Predictive Model}
\label{sig_model}

We used a rolling window analysis to learn temporal dependencies in the data. This is a commonly used method to estimate changes and reveal patterns in sequential data over time \cite{zivot2007modeling}. The rolling window analysis allows to estimate the variability of data within certain time intervals and to compare between them. 

The proposed predictive model is based on logistic regression, which learns patterns in data within the window of size k-weeks to predict the future occurrence of an episode at one-week time horizon. Two linear logistic models were trained independently to predict depressive and manic episodes. Using the definition of depressive and manic episodes, we developed an algorithm which identifies episodes in patients’ data and marks the weekly observations with a binary label (0 or 1) corresponding to the absence or presence of an episode. 

The features used in logistic regression to predict outcomes, were extracted using the signature method \cite{chevyrev2016primer}. The novel signature method, from stochastic analysis, has the ability to capture important properties of complex time-ordered data. It maps N unique data streams into a single piece-wise continuous path in N dimensions with the iterated integrals of such path representing a feature set (signature features) of the data. The signature is an infinite sequence of ordered iterated integrals and in practice. For machine learning applications and data analyses, we truncate the infinite sequence at some level $L$ by taking only first terms and use them as features. 

The signature method allows missing responses to be incorporated into the analysis by introducing a new indicator binary variable, which takes values 1 or 0, indicates whether a response is missing or present respectively \cite{little2014statistical}. The indicator variable stream is then combined with N data streams of interest and mapped into a lifted path in N+1 dimensions. Practically, we used the self-reported data streams (QIDS and Altman questionnaires) together with their missing responses indicator data streams. The detailed description, rigorous mathematical foundations and algorithms of the signature methods are beyond the scope of the current paper and have been extensively covered in \cite{kormilitzin2016application, levin2013learning, gyurko2013extracting, yang2015chinese, lyons2014rough, kiraly2016kernels, yin2013icdar, graham2013sparse, chen1957integration, lyons2011inversion, lyons2007differential, hambly2010uniqueness, hairer2014theory, yang2016rotation, xie2016learning, lai2017toward} and references therein.

In order to estimate the generalisation error of predictive models, we split the entire data set into two non-overlapping groups of patients for training (67$\%$) and testing (33$\%$) data sets, while preserving the ratio of weeks in episodes to weeks without episodes in both sets. The stratified splitting strategy allowed the variance and bias of a model to be minimised \cite{kohavi1995study, cawley2010over}. A model was trained and cross-validated using the data only from one group and then the data from the withheld second group of patients was used to assess the classification metrics. The process of random splitting of patients into two disjoint groups was repeated 100 times to estimates the uncertainty of the metrics.

The elastic net regularisation scheme \cite{zou2005regularization} was used to prevent the models from overfitting the data. The elastic net regularisation introduces the convex combination of L1 and L2 penalties, mitigating the problem of multicollinearity of features and improving the stability of a classifier. As an additional preprocessing step, the signature features matrix was standardised along columns to have zero mean and unit variance. To account for the variability in longitudinal data, we used the lead-lag transformation, where the lead transform of the signal is paired with its lagged transform of the signal. To assess the performance of the classification procedure we computed accuracy, sensitivity, specificity, and positive predicted value (PPV). Additionally, we used the area under the receiver operating characteristic curve (AUC) to assess the performance of the classification models at different values of a threshold. 

We compared the signature-based model (Sig) to four baseline models, where three of them based only on one of the mean value (Mean), variability (Rmssd) and a number of missing responses (MissRes) within the window predictors and the fourth model comprises all these three predictors simultaneously in a linear combination (MRM). The variability was computed, ignoring the missing responses, using the root mean square successive distance (RMSSD) \cite{Electrophysiology1043}, which captures the temporal dependencies in the data. 

\subsection{Parameters of the Model}

The longitudinal self-reported data were partitioned into k-week intervals, transformed into a set of features using the signature method and the resulting features were used as inputs to logistic regression to predict a binary outcome, presence or absence of an episodes, at the week following the interval. The intervals that precede the episodes (precursors to episodes) are labelled as positives (‘1’). The data from QIDS and AMSR questionnaires were used to train models predicting respectively depressive and manic episodes. Intervals that precede missing responses or contain only one non-missing response were excluded. 
Additionally, we sought to optimise the classification model using the hyperparameters of the penalised logistic regression during the 10-fold cross-validation phase with the parameter $\lambda \in \{0,0.1,0.2,0.3,...,1.0\}$ that mixes L1 and L2 penalties.

We used the Python Pandas package (version 0.20.1) \cite{mckinney2010data} for statistical analysis, data manipulations and processing, Python Scikit-learn package (version 0.18.1) \cite{pedregosa2011scikit} for implementing machine learning tasks and Matplotlib for plotting and graphics (version 2.0.1) \cite{hunter2007matplotlib}.  

\section{Results}

\subsection{Predicting Depressive and Manic Episodes}

\begin{figure*}[]
  \centering
  \subfigure[]{\includegraphics[scale=0.38]{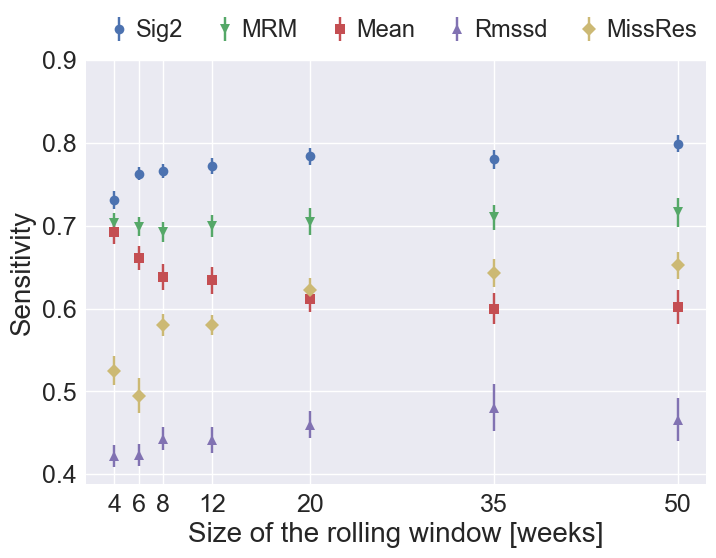}}\quad
  \subfigure[]{\includegraphics[scale=0.38]{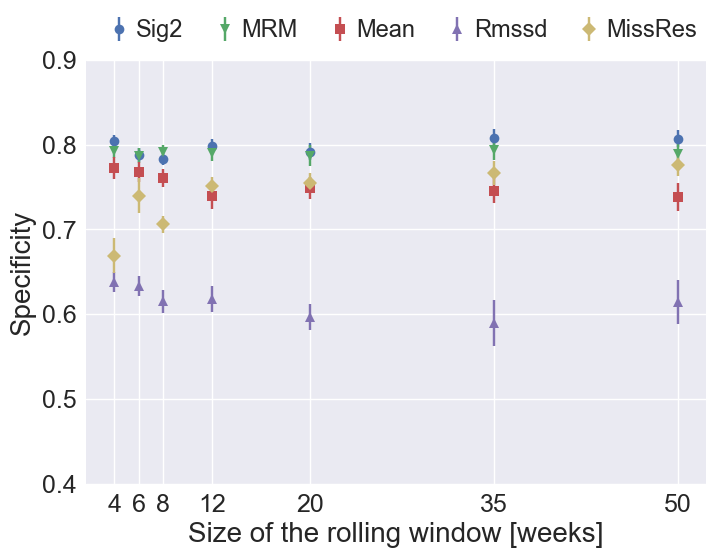}}\quad
  \subfigure[]{\includegraphics[scale=0.38]{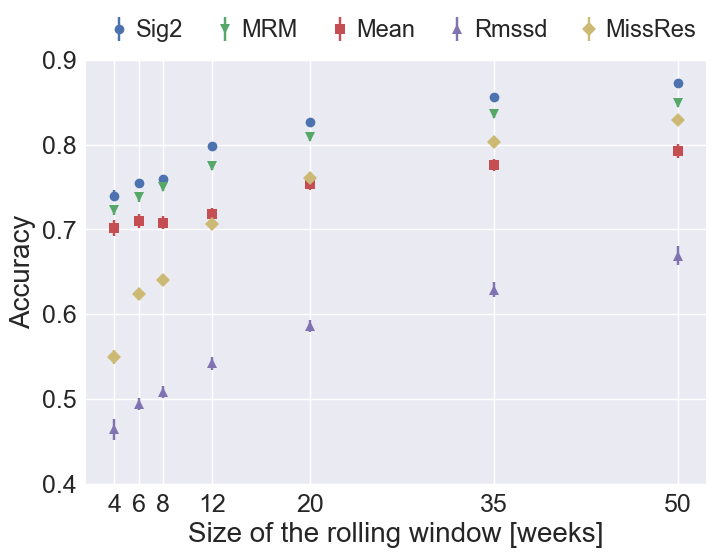}}\quad
  \subfigure[]{\includegraphics[scale=0.38]{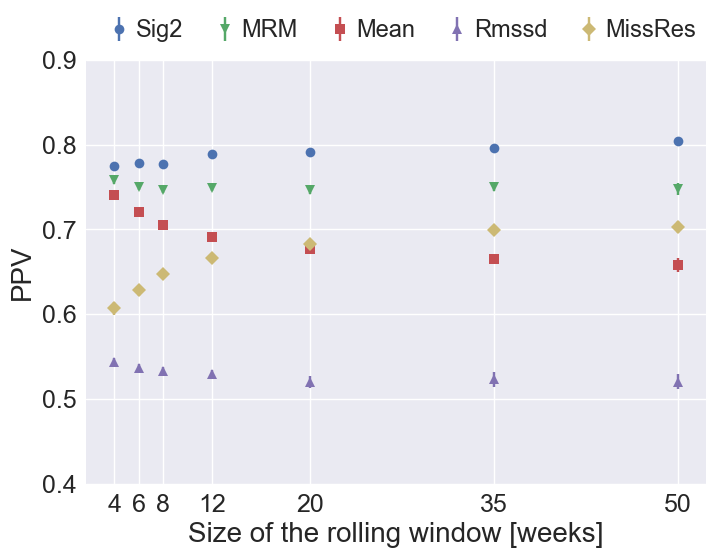}}\quad
  \subfigure[]{\includegraphics[scale=0.38]{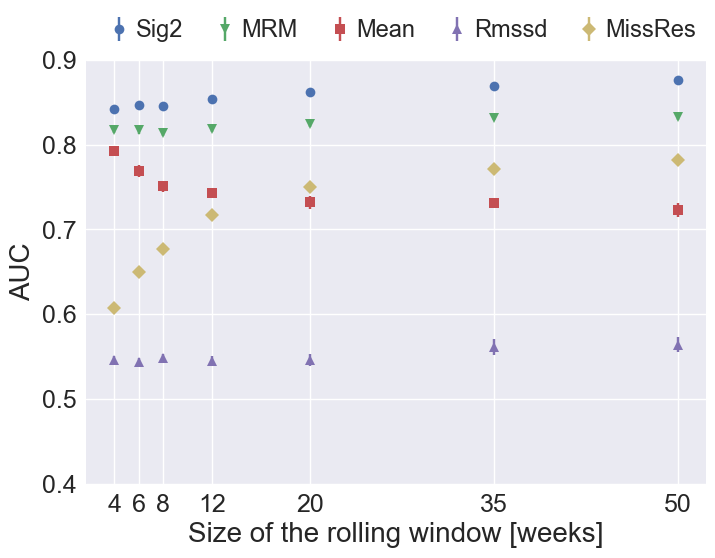}}
  \caption{Classification metrics for depression model presented as mean with 95$\%$CI for different size of the rolling window.}
  \label{fig:depres_metrics}
\end{figure*}

\begin{figure*}[]
  \centering
  \subfigure[]{\includegraphics[scale=0.38]{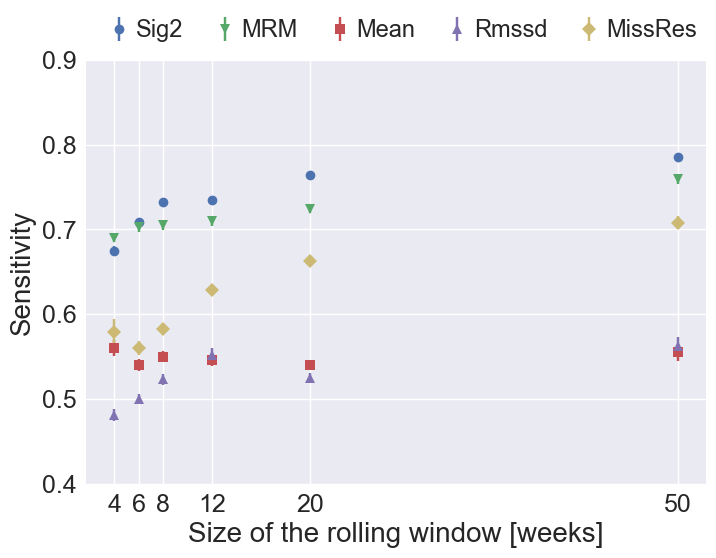}}\quad
  \subfigure[]{\includegraphics[scale=0.38]{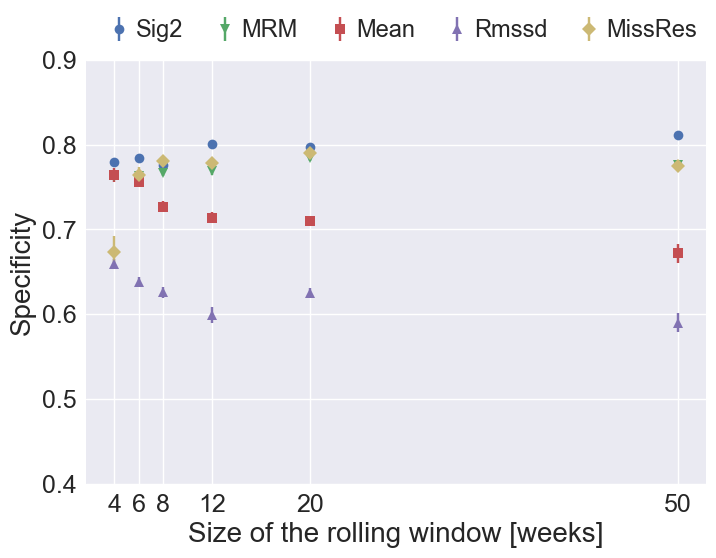}}\quad
  \subfigure[]{\includegraphics[scale=0.38]{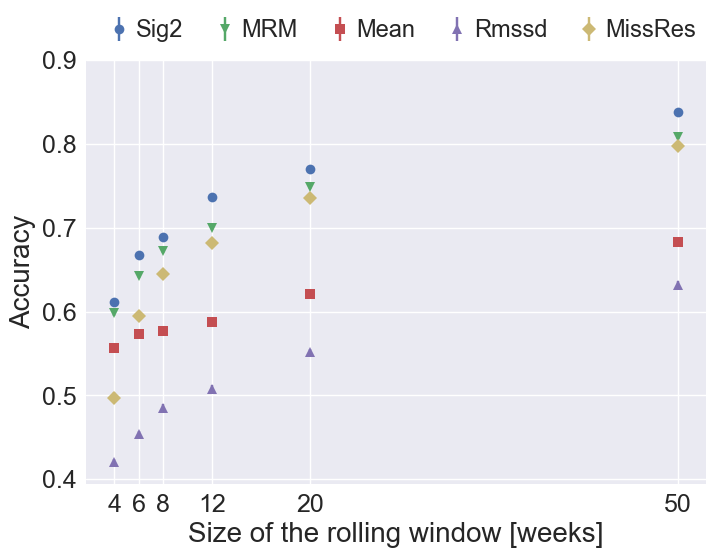}}\quad
  \subfigure[]{\includegraphics[scale=0.38]{plots/PPV_q}}\quad
  \subfigure[]{\includegraphics[scale=0.38]{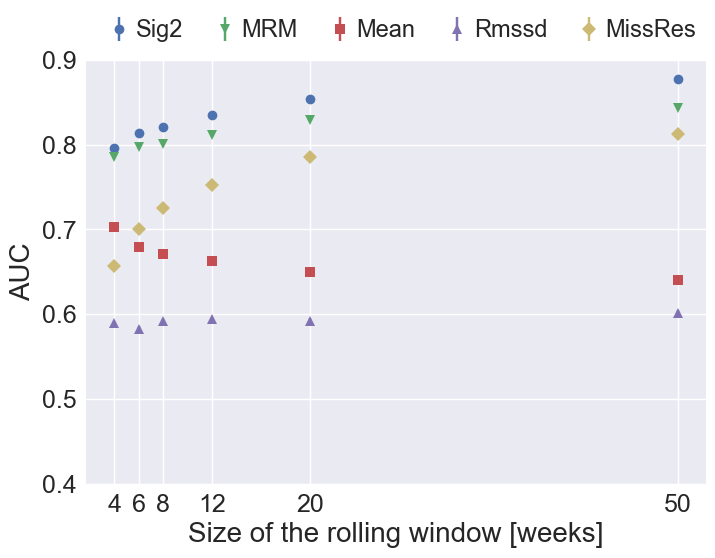}}
  \caption{Classification metrics for mania model presented as mean with 95$\%$CI for different size of the rolling window.}
  \label{fig:mania_metrics}
\end{figure*}

The performance of the signature-based model and comparison to the baseline models for depression and mania trained on corresponding test sets are summarised in Tables \ref{table:deprepres_results}-\ref{table:mania_results} respectively. Classification results evaluated on the corresponding test sets of the models for depression and mania using 100 repetition of random splitting to train/test (67$\%$,/33$\%$) sets. The features used in the model are elements of a truncated signature at level 2. The repeated results of metrics were nearly normal distributed and following the suggested procedure \cite{salzberg1997comparing, dietterich1998approximate}, we used the corrected paired t-test to pair-wise compare the performance of all metrics of five models. We found a strong statistical significance of their differences (p-value $<$ 0.001), also for depression and mania models. The stability of the classification metrics was estimated by repeating the training and testing the models 100 times and the results presented as mean and standard deviation. The classification metrics of models for depression and mania as function of the size of the rolling window are presented in Figures \ref{fig:depres_metrics}-\ref{fig:mania_metrics}. 

\subsection{Experimental Clinical Applications}

The proposed models for classification of intervals are potentially useful for clinical insights and for identifying particular states, where a patient is in transition between the episodes. First, we are interested in examining whether it is possible to distinguish between the intervals which are close to an episode and those which are further away. For the sake of simplicity and demonstration of the conceptual approach, we consider intervals of fixed length of size 6 weeks. Two types of intervals are: a ‘wellness’ and a ‘precursor’ interval, which are spaced 14 and $n$ weeks prior to the beginning of an episode respectively. By varying $n = \{0,1,2,3,4,6\}$, and applying the developed models, we will measure the area under the ROC curve (AUC) to assess the classification performance. We restrict the wellness and the precursor intervals from overlapping. For classification procedure we considered only the signature-based model. To estimate the variability due to split of the data into training and testing sets (67$\%$ / 33$\%$) we repeated the process 100 times. The results are presented as mean and standard deviation. The AUC estimations are summarised in Table  \ref{table:experiment}. 

\section{Discussion}

\subsection{Principal Results}

In this study we have presented an application of the signature method for modelling and prediction of mood episodes in bipolar disorder. We demonstrated that the results of the signature-based model outperform four other models based on the manually selected predictors. The main advantage of the signature method is that it allows to systematically combine multimodal longitudinal data streams, including the distribution of missing responses, and to extract features used in predictive models, avoiding a complicated feature engineering process. The signature terms faithfully represent the underlying data, have an interpretable geometrical meaning as functions of data and should be used as canonical features in machine learning and data analysis tasks. Intuitively, one can think of the signature terms as of an ordered collection of the sample statistical moments. The lead-lag transformation naturally captures the successive variability in the data, avoiding the need for designing a special feature for that task as we did with the RMSSD.

The plot of the classification metrics (Figures \ref{fig:depres_metrics}-\ref{fig:mania_metrics}) demonstrate the dependence of the predictive performance of the models on the size of the rolling window. For small values of the window size (k $<$ 12) all models perform similar, while the signature-based model, MRM and MissRes, continue improving as the number of historical observations grows and the Mean and Rmssd models decline or saturate. The predictive performance does ultimately depend on the ability of models to account for the temporal dependencies. The signature model is based on the unique set of predictors derived from a trajectory (‘a path’) of the longitudinal self-reported responses, where the predictors account for the sequential dependencies through the path integrals and thus more observations lead to a higher accuracy of predictions. The Mean model, in contrast with the others, strongly declines when used with a large number of historical observations. That manifests the fact, that the average score over a long interval of observations is a poor predictor of future episodes, because it does not take into account the temporal dependencies. Interestingly, it appears that the variability of the rating scale along is a very poor predictor of future episodes, even though it does account for the sequential dependencies in the data through the RMSSD metric. Another interesting observation is that the number of missing responses along within the window can predict with up to 80$\%$ accuracy the future episodes of depression and mania. Finally, we observed that the composite model (MRM) that comprises all three predictors linearly, underperforms the accuracy of the signature-based model on average by 2.36$\%$.

The developed models aim to discover the temporal patterns of mood episodes and accurately predict the possible future outcomes. The proposed algorithms can be potentially deployed in modern health-monitoring platforms and serve as a patient self-management tool. Detection of precursors to upcoming mood episodes and early intervention of the healthcare professionals can help to reduce the severity of symptoms in patients with bipolar disorder. However, the developed predictive models do not take into account the unique personal traits of patients, for example, the distribution of missing values and the item responses. The main problem of training an individual model is due to relatively small number of episodes each participant experiences during the self-monitoring. 

While the results of the current work have demonstrated the feasibility of building the predictive models based on the signature transformation of the self-reported data alone, more in-depth research is needed to refine the model. The current analysis considered only a relatively small number of patients who satisfied the inclusion criteria and in order to develop robust models, in the future we are planning to replicated the proposed method using a larger cohort. It has been shown [40] that training a model on a selected subgroup of patients who share common traits leads to a significant improvement of accuracy of the predictive models. We are planning to address the problem of clustering patient in our future works and retrain the predictive models for each cluster.

Apart from the identifying early signs of deterioration, it is also clinically interesting to identify early signs of improvement. The approach to find precursors to wellness is conceptually similar to the one we presented in this work, but requires different labelling of intervals and retraining the models. 

The developed approach to interval classification allows us potentially to identify the early signs of transitions between the mood states in patients. We tested whether the we can distinguish between an interval which is not in an episode and an interval which is n-weeks close to the beginning of an episode. The results presented in Table \ref{table:experiment} indicate that intervals closer to an impeding are distinct than those which are far apart. However, the choice of the interval size (6-weeks) and the subjective definition of the wellness intervals (14 weeks prior to an episode) may not be optimal to conclude the results. Further investigation needed.

In this work we used only linear models for prediction of future episodes. Linear models preferable over more complex and non-linear ones, especially with medical data, as they allow to understand clearly which variables used as predictors and their influence on the dependent variable. However, the important clinical condition which has not yet been addressed through our research is the case of mixed episodes, where both depression and mania may occur simultaneously \cite{vieta2013mixed}. The signature method allows to combine and develop a model using both QIDS and AMSR data streams to predict the future states including the missing ones. We are planning to address this problem in a future work.

\subsection{Limitations}

Notwithstanding the signature approach outperforms other models based on the manually created predictors, the full potential of the signature method, which has been shown to be superior to state-of-the-art methods in other fields \cite{yang2017leveraging} has not yet been demonstrated on the longitudinal mood data. We refer the limited power of the proposed method to the challenging nature of the self-reported longitudinal data, e.g. presence of non-randomly distributed missing values and more importantly, to the diversity of the cohort and lack of regularly time-stamped records of exogenous factors (for example, medications and hospitalisations). The moderate accuracies of the models are also influenced by the assumption that all patients share the same patterns of early signs of deterioration, which might not be true. That assumption allowed us to train and test the models on randomly selected non-overlapping sets of patients from the entire cohort. With more data collected over longer period of the self-monitoring, we are interested in construction of individual models, to examine the diversity in patients. Additionally, the proposed method for incorporating of missing responses might not be optimal. In this work we laid a significant groundwork with a lot of open questions for further research and are planning to address the aforementioned problems in subsequent publications.

\subsection{Conclusions}

In this work we developed and compared several linear models for predicting the episodes of depression and mania in patients with bipolar disorders. We found that the accuracy of predictions depends on the length of historical observations (the size of the rolling window). The accuracy of the signature-based model is higher than all others by the amount of at least as 2.36$\%$ and 3.37$\%$ for depression and mania respectively. The signature method represents a systematic approach for feature extraction from and modelling functions on streams of data. We found that the linear combination of three predictors (Mean, Rmssd and MissRev) outperforms each of them individually for predicting the future episodes in both models for depression and mania.

\section*{Ethics}

The study was approved by Oxfordshire Research Ethics Committee A (reference no: 10/H0604/13). All participants in gave written, informed consent.

\section*{Data Accessibility}
Data can be requested from JRG

\section*{Competing Interests}
PJH, AK, TJL, KEAS and JRG declare no competing interests. 

\section*{Author’s Contributions}
JRG designed the trial. JRG coordinated the study. AK and TJL conducted the analysis. AK, TJL and KEAS drafted the paper, which was reviewed by all authors. 

\section*{Acknowledgements} 
We thank Maarten De Vos, Guy Goodwin, Keltie McDonald, Nick Palmius and Athanasios Tsanas for valuable discussions. 

\section*{Funding}
AK, TJL, KEAS, JRG and PJH are supported by a Wellcome Trust Strategic Award CONBRIO: Collaborative Oxford Network for Bipolar Research to Improve Outcomes, Reference number 102616/Z.  JRG, PH and KEAS are supported by the NIHR Oxford Health Biomedical Research Centre. TJL acknowledges the support of NCEO project NERC, ERC grant number 291244, EPSRC grant number EP/H000100/1 and by the Alan Turin Institute under the EPSRC grant EP/N510129/1. No funder had any role in the study design; data collection, analysis, or interpretation of data; writing of the report; or in the decision to submit the paper for publication. The views expressed are those of the authors and not necessarily those of the NHS, the NIHR or the Department of Health.

\bibliography{sig_mod}
\bibliographystyle{icml2017}

\begin{table*}
\centering
\begin{tabular}{l|r|r|r|r}\hline
\textbf{Diagnosis}  &\multicolumn{2}{c|}{BP-I (N=148)} &\multicolumn{2}{c}{BP-II (N=113)} \\\hline\hline
\textbf{Sex}        & Male & Female & Male & Female \\
                    & 51 (34$\%$)  & 97 (66$\%$) & 39 (35$\%$) & 74 (65$\%$) \\\hline
\textbf{Ethnicity}  &  &  &  & \\
\multicolumn{1}{r|}{White}       & 50 $(98\%)$  &  86 ($89\%$) & 35 ($92\%$) & 66 ($92\%$)\\\hline
\textbf{Age}  &  &  &  & \\
\multicolumn{1}{r|}{years}       & & &  & \\
\multicolumn{1}{r|}{17 - 25}    & 4  (0.08$\%$) & 12 (0.13$\%$) & 7  (0.18$\%$)  & 13  (0.18$\%$) \\
\multicolumn{1}{r|}{25 - 35}    & 13 (0.25$\%$) & 25 (0.27$\%$) & 7  (0.18$\%$)  & 23  (0.32$\%$) \\
\multicolumn{1}{r|}{35 - 45}    & 16 (0.31$\%$) & 32 (0.34$\%$) & 9  (0.23$\%$)  & 13  (0.18$\%$) \\
\multicolumn{1}{r|}{45 - 55}    & 10 (0.20$\%$) & 14 (0.15$\%$) & 9  (0.23$\%$)  & 12  (0.16$\%$) \\
\multicolumn{1}{r|}{55 - 65}    & 6  (0.12$\%$) & 6  (0.06$\%$) & 7  (0.18$\%$)  & 7   (0.10$\%$) \\
\multicolumn{1}{r|}{65 - 75}    & 2  (0.04$\%$) & 5  (0.05$\%$) & 0     (0$\%$)  & 5   (0.09$\%$) \\
\multicolumn{1}{r|}{Mean (SEM)} & 41.70 (1.77) & 38.70 (1.36)& 39.98 (2.23) & 38.69 (1.77)\\
\multicolumn{1}{r|}{Range}      & 17 - 70      & 17 - 74     & 20 - 64      & 17 - 72.5\\
\multicolumn{1}{r|}{Median (IQR)}& 42 (16)     & 38 (18)     & 37 (22.5)    & 35 (23.5)\\\hline
\textbf{Education}  &  &  &  & \\
\multicolumn{1}{r|}{Tertiary}       & 32 (0.63$\%$) & 52 (0.53$\%$) & 18 (46$\%$) & 44 ($60\%$)\\\hline
\textbf{Adherence}  &  &  &  & \\
\multicolumn{1}{r|}{Mean (SEM)}                  & 0.45 (0.04) & 0.38 (0.05) & 0.41 (0.05) & 0.41 (0.04)\\
\multicolumn{1}{r|}{Adjusted (Mean (SEM))}       & 0.64 (0.02) & 0.59 (0.03) & 0.61 (0.02) & 0.65 (0.04)\\\hline
\textbf{Episodes}  &  &  &  & \\
\multicolumn{1}{r|}{Depression}  & 23 (45$\%$) & 58 (60$\%$) & 17 (46$\%$) & 57 (77$\%$)\\
\multicolumn{1}{r|}{Mania}       & 33 (65$\%$) & 68 (70$\%$) & 26 (67$\%$) & 54 (73$\%$)\\\hline
\end{tabular}
\caption{Demographic summary of the data used for modelling.}
\label{table:demographics_results}
\end{table*}

\begin{table*}
\centering
\begin{tabular}{l|r|r|r|r|r|r}\hline
\textbf{weeks} & \textbf{Model} & \textbf{Sensitivity} & \textbf{Specificity} & \textbf{Accuracy} & \textbf{PPV} & \textbf{AUC} \\\hline\hline
 & Sig & \textbf{0.731}(0.054) & \textbf{0.804}(0.037) & \textbf{0.741}(0.032) & \textbf{0.775}(0.019) & \textbf{0.842}(0.021) \\
 & MRM & 0.703(0.062) & 0.793(0.045) & 0.723(0.032) & 0.758(0.021) & 0.818(0.023) \\
k=4 & Mean & 0.692(0.071) & 0.773(0.067) & 0.702(0.048) & 0.741(0.024) & 0.793(0.033) \\
 & Rmssd & 0.422(0.068) & 0.638(0.063) & 0.464(0.061) & 0.544(0.022) & 0.546(0.024) \\
 & MissRes & 0.525(0.089) & 0.669(0.106) & 0.55(0.042) & 0.607(0.038) & 0.608(0.024) \\\hline
 & Sig & \textbf{0.763}(0.042) & \textbf{0.788}(0.041) & \textbf{0.755}(0.025) & \textbf{0.779}(0.017) & \textbf{0.847}(0.017) \\
 & MRM & 0.699(0.06) & 0.787(0.046) & 0.738(0.028) & 0.75(0.02) & 0.817(0.021) \\
k=6 & Mean & 0.661(0.074) & 0.768(0.059) & 0.71(0.043) & 0.721(0.024) & 0.769(0.034) \\
 & Rmssd & 0.423(0.066) & 0.633(0.059) & 0.494(0.037) & 0.537(0.024) & 0.543(0.028) \\
 & MissRes & 0.495(0.105) & 0.74(0.106) & 0.624(0.038) & 0.628(0.035) & 0.65(0.026) \\\hline
 & Sig & \textbf{0.766}(0.042) & 0.783(0.036) & \textbf{0.76}(0.025) & \textbf{0.777}(0.017) & \textbf{0.846}(0.016) \\
 & MRM & 0.692(0.062) & \textbf{0.791}(0.044) & 0.75(0.026) & 0.747(0.02) & 0.814(0.019) \\
k=8 & Mean & 0.638(0.078) & 0.761(0.054) & 0.708(0.039) & 0.705(0.022) & 0.751(0.033) \\
 & Rmssd & 0.443(0.072) & 0.615(0.068) & 0.508(0.036) & 0.533(0.024) & 0.548(0.027) \\
 & MissRes & 0.58(0.07) & 0.706(0.051) & 0.64(0.028) & 0.647(0.023) & 0.677(0.024) \\\hline
 & Sig & \textbf{0.772}(0.048) & \textbf{0.798}(0.046) & \textbf{0.799}(0.019) & \textbf{0.789}(0.014) & \textbf{0.854}(0.014) \\
 & MRM & 0.7(0.067) & 0.79(0.049) & 0.775(0.024) & 0.749(0.018) & 0.819(0.017) \\
k=12 & Mean & 0.634(0.085) & 0.739(0.078) & 0.718(0.039) & 0.691(0.022) & 0.743(0.032) \\
 & Rmssd & 0.441(0.08) & 0.618(0.08) & 0.542(0.04) & 0.529(0.029) & 0.545(0.031) \\
 & MissRes & 0.58(0.063) & 0.751(0.057) & 0.707(0.035) & 0.665(0.027) & 0.717(0.025) \\\hline
 & Sig & \textbf{0.784}(0.053) & \textbf{0.792}(0.049) & \textbf{0.827}(0.018) & \textbf{0.792}(0.018) & \textbf{0.862}(0.014) \\
 & MRM & 0.705(0.085) & 0.787(0.064) & 0.809(0.023) & 0.745(0.025) & 0.825(0.018) \\
k=20 & Mean & 0.612(0.085) & 0.749(0.068) & 0.754(0.038) & 0.677(0.031) & 0.732(0.037) \\
 & Rmssd & 0.46(0.084) & 0.597(0.078) & 0.586(0.038) & 0.52 (0.035) & 0.546(0.038) \\
 & MissRes & 0.622(0.076) & 0.755(0.059) & 0.761(0.033) & 0.683(0.027) & 0.75(0.027) \\\hline
 & Sig & \textbf{0.799}(0.052) & \textbf{0.807}(0.051) & \textbf{0.873}(0.018) & \textbf{0.805}(0.022) & \textbf{0.877}(0.013) \\
 & MRM & 0.716(0.088) & 0.789(0.063) & 0.849(0.023) & 0.748(0.037) & 0.833(0.019) \\
k=50 & Mean & 0.602(0.104) & 0.738(0.084) & 0.79 (0.042) & 0.658(0.041) & 0.723(0.04) \\
 & Rmssd & 0.466(0.135) & 0.614(0.132) & 0.669(0.058) & 0.52(0.045) & 0.564(0.045) \\
 & MissRes & 0.652(0.081) & 0.776(0.064) & 0.829(0.029) & 0.703(0.035) & 0.782(0.03) \\\hline
\end{tabular}
\caption{Prediction results for depression on the test set of the signature-based and four baseline models for different size of the rolling window. The results are presented as mean(SD).}
\label{table:deprepres_results}
\end{table*}

\begin{table*}
\centering
\begin{tabular}{l|r|r|r|r|r|r}\hline
\textbf{weeks} & \textbf{Model} & \textbf{Sensitivity} & \textbf{Specificity} & \textbf{Accuracy} & \textbf{PPV} & \textbf{AUC} \\\hline\hline
 & Sig & \textbf{0.675}(0.05) & \textbf{0.78}(0.035) & \textbf{0.611}(0.031) & \textbf{0.747}(0.020) & \textbf{0.796}(0.021) \\
 & MRM & 0.69(0.047) & 0.763(0.037) & 0.598(0.031) & 0.741(0.021) & 0.786(0.021) \\
k=4 & Mean & 0.560(0.091) & 0.764(0.085) & 0.556(0.059) & 0.697(0.038) & 0.703(0.033) \\
 & Rmssd & 0.481(0.072) & 0.659(0.06) & 0.42(0.04) & 0.598(0.024) & 0.590(0.023) \\
 & MissRes & 0.579(0.149) & 0.674(0.189) & 0.497(0.065) & 0.644(0.079) & 0.657(0.026) \\\hline
 & Sig & \textbf{0.709}(0.049) & \textbf{0.785}(0.043) & \textbf{0.667}(0.026) & \textbf{0.759}(0.019) & \textbf{0.814}(0.018) \\
 & MRM & 0.703(0.054) & 0.763(0.045) & 0.643(0.027) & 0.744(0.021) & 0.797(0.021) \\
k=6 & Mean & 0.54(0.066) & 0.756(0.05) & 0.573(0.044) & 0.677(0.027) & 0.679(0.035) \\
 & Rmssd & 0.5(0.058) & 0.638(0.056) & 0.454(0.037) & 0.586(0.025) & 0.583(0.029) \\
 & MissRes & 0.56(0.083) & 0.764(0.094) & 0.595(0.04) & 0.69(0.039) & 0.701(0.022) \\\hline
 & Sig & \textbf{0.732}(0.047) & 0.776(0.045) & \textbf{0.689}(0.024) & \textbf{0.761}(0.017) & \textbf{0.821}(0.014) \\
 & MRM & 0.705(0.053) & 0.767(0.043) & 0.672(0.025) & 0.745(0.017) & 0.801(0.016) \\
k=8 & Mean & 0.55(0.069) & 0.727(0.061) & 0.577(0.042) & 0.658(0.024) & 0.671(0.03) \\
 & Rmssd & 0.523(0.065) & 0.626(0.064) & 0.485(0.042) & 0.584(0.025) & 0.592(0.029) \\
 & MissRes & 0.583(0.067) & \textbf{0.781}(0.056) & 0.645(0.04) & 0.703(0.024) & 0.725(0.023) \\\hline
 & Sig & \textbf{0.735}(0.043) & \textbf{0.801}(0.045) & \textbf{0.737}(0.018) & \textbf{0.776}(0.016) & \textbf{0.835}(0.014) \\
 & MRM & 0.71(0.054) & 0.769(0.051) & 0.7 (0.023) & 0.748(0.018) & 0.811(0.017) \\
k=12 & Mean & 0.546(0.07) & 0.714(0.061) & 0.588(0.042) & 0.644(0.024) & 0.663(0.033) \\
 & Rmssd & 0.552(0.08) & 0.599(0.091) & 0.508(0.041) & 0.581(0.026) & 0.594(0.028) \\
 & MissRes & 0.628(0.052) & 0.778(0.051) & 0.682(0.026) & 0.717(0.023) & 0.753(0.02) \\\hline
 & Sig & \textbf{0.764}(0.039) & \textbf{0.797}(0.043) & \textbf{0.77}(0.023) & \textbf{0.784}(0.016) & \textbf{0.854}(0.013) \\
 & MRM & 0.724(0.045) & 0.785(0.042) & 0.749(0.022) & 0.759(0.016) & 0.829(0.014) \\
k=20 & Mean & 0.54(0.063) & 0.710(0.052) & 0.621(0.036) & 0.632(0.024) & 0.65(0.032) \\
 & Rmssd & 0.525(0.059) & 0.625(0.059) & 0.552(0.039) & 0.58 (0.026) & 0.592(0.035) \\
 & MissRes & 0.663(0.05) & 0.79(0.041) & 0.736(0.026) & 0.733(0.018) & 0.786(0.019) \\\hline
 & Sig & \textbf{0.786}(0.046) & \textbf{0.811}(0.04) & \textbf{0.838}(0.022) & \textbf{0.8}(0.019) & \textbf{0.878}(0.013) \\
 & MRM & 0.76(0.057) & 0.776(0.05) & 0.808(0.026) & 0.77(0.021) & 0.843(0.016) \\
k=50 & Mean & 0.555(0.1) & 0.672(0.114) & 0.683(0.052) & 0.608(0.031) & 0.64(0.036) \\
 & Rmssd & 0.563(0.101) & 0.59(0.116) & 0.632(0.043) & 0.577(0.027) & 0.602(0.035) \\
 & MissRes & 0.708(0.074) & 0.775(0.065) & 0.798(0.036) & 0.741(0.025) & 0.813(0.024) \\\hline
\end{tabular}
\caption{Prediction results for mania on the test set of the signature-based and four baseline models for different size of the rolling window. The results are presented as mean(SD).}
\label{table:mania_results}
\end{table*}

\begin{table*}
\centering
\begin{tabular}{c|c|c}\hline
\textbf{weeks before episode} & \textbf{Depression} & \textbf{Mania} \\ \hline\hline
n = 0 & 0.696(0.048) & 0.742(0.051) \\
n = 1 & 0.676(0.059) & 0.675(0.074) \\
n = 2 & 0.606(0.063) & 0.637(0.038) \\
n = 3 & 0.634(0.061) & 0.612(0.063) \\
n = 4 & 0.614(0.078) & 0.558(0.048) \\
n = 6 & 0.637(0.066) & 0.571(0.071) \\\hline
\end{tabular}
\caption{The results of discrimination between the wellness and the precursor intervals of 6 weeks. Precursor intervals are tested at n week before an impending an episode. The values are the area under the ROC curve (AUC) and presented as mean(SD).}
\label{table:experiment}
\end{table*}



\end{document}